\pdfoutput=1

\documentclass[11pt]{article}

\usepackage[final]{ACL2023}

\usepackage{times} \usepackage{latexsym} \usepackage{amsmath} \usepackage{tikz}
\usepackage{amssymb} \usepackage{hyperref} \usepackage{graphicx}
\usepackage{titlesec}
\usepackage{hyperref}
\usepackage{cprotect}
\usepackage{float}

\titleformat*{\section}{\Large\bfseries}
\titleformat*{\subsection}{\large\bfseries}
\titleformat*{\paragraph}{\normalsize\bfseries}

\usepackage[T1]{fontenc}

\usepackage[utf8]{inputenc}

\usepackage{microtype}

\usepackage{inconsolata}

\renewcommand{\vec}[1]{\mathbf{#1}}
\title{Comparing and combining some popular NER approaches on Biomedical tasks}

\author{Harsh Verma,  Sabine Bergler,  Narjesossadat Tahaei \\
        CLaC Labs, Concordia University \\
        \{h\_ver, bergler, n\_tahaei\} @cse.concordia.ca}

\begin{document} 
\maketitle 
\begin{abstract}
We compare three simple and popular approaches for NER: 1) \texttt{SEQ} (sequence-labeling with a linear token classifier) 2) \texttt{SeqCRF} (sequence-labeling with Conditional Random Fields), and 3) \texttt{SpanPred} (span-prediction with boundary token embeddings). 
We compare the approaches on 4 biomedical NER tasks: GENIA, NCBI-Disease, LivingNER (Spanish), SocialDisNER (Spanish).
The \texttt{SpanPred} model demonstrates state-of-the-art performance on LivingNER and SocialDisNER, 
improving F1 by 1.3 and 0.6 F1 respectively. 
The \texttt{SeqCRF} model also demonstrates state-of-the-art performance on LivingNER and SocialDisNER, improving F1 by 0.2 F1 and 0.7 respectively.
The \texttt{SEQ} model is competitive with the state-of-the-art on the LivingNER dataset.
We explore some simple ways of combining the three approaches. 
We find that majority voting consistently gives high precision and high F1 across all 4 datasets.
Lastly, we implement a system that learns to combine the predictions of \texttt{SEQ} and \texttt{SpanPred},
generating systems that consistently give high recall and high F1 across all 4 datasets. 
On the GENIA dataset, we find that our learned combiner system significantly 
boosts F1(+1.2) and  recall(+2.1) over the systems being combined. 
We release all the well-documented code necessary to reproduce all systems at  \href{https://github.com/flyingmothman/bionlp}{this Github repository}.
\end{abstract}

\section{Introduction} 
NER has frequently been formulated as a sequence-labeling 
problem  \cite{chiu-nichols-2016-named, ma-hovy-2016-end, wang-etal-2022-damo} 
in which a model learns to label each token using a labeling scheme such as 
BIO(\textit{beginning}, \textit{inside},
\textit{outside}). However, in recent years people have also formulated
the NER task as a span-prediction problem
\cite{jiang-etal-2020-generalizing, li-etal-2020-unified, spanner, zhang2023optimizing}
where spans of text are represented and labeled with entity types. \par 
Let \texttt{SEQ} be the simplest sequence-labeling model
which represents each token using a language model and then 
classifies each token-representation with a linear layer. 
Let \texttt{SeqCRF} be another popular sequence-labeling model which is identical 
to \texttt{SEQ} model except that
the token representations from the language model are fed into a
linear-chain conditional random field layer\cite{lafferty2001conditional, lample-etal-2016-neural}. 
Let \texttt{SpanPred}\cite{lee-etal-2017-end, jiang-etal-2020-generalizing} be a 
model that represents every possible span of text using 
two token-embeddings located at the its boundary, 
and then classifies every span-representation using a linear layer. We describe all three models
in detail in \autoref{models}. 
We evaluate \texttt{SEQ}, \texttt{SeqCRF}, and \texttt{SpanPred} models 
on four biomedical NER tasks: GENIA\cite{kim2003genia}, NCBI-Disease\cite{dougan2014ncbi}, 
LivingNER(Spanish)\cite{miranda2022mention}, and SocialDisNER(Spanish)\cite{gasco-sanchez-etal-2022-socialdisner}. 
Despite being simple, the \texttt{SpanPred} and \texttt{CRF} models 
improve the state-of-the-art on the LivingNER and SocialDisNER tasks.\par

\cite{spanner} show that the sequence-labeling approaches(eg. \texttt{Seq} and \texttt{SeqCRF}) and span-prediction approaches(eg. \texttt{SpanPred}) have \textit{different} strengths and weaknesses \textit{while} having similar(F1) performance. This motivated us to try and combine \texttt{Seq},  \texttt{SeqCRF}, and \texttt{SpanPred} models using two simple methods and study the results. We refer to the two simple methods as \texttt{Union} and \texttt{MajVote}. \texttt{Union} is inspired by the set(mathematical) union operation and it simply involves "unioning" the sets of predictions made by the models. \texttt{MajVote} is the classic majority voting method. We find that \texttt{MajVote} can yield systems that have both high precision and high F1.

\par Inspired by the boost in recall(and the corresponding drop in precision) resulting from the \texttt{Union} method, we implemented a combiner system
(which we refer to as \texttt{Meta}) that aims to \textit{combat} the drop in precision as a result of the \texttt{Union} method. We find that \texttt{Meta} shows very promising signs of increasing precision while preserving high recall and high F1. \texttt{Meta} borrows ideas from work on generating span representations using "solid markers"\cite{baldini-soares-etal-2019-matching, xiao-etal-2020-denoising, ye-etal-2022-packed}, work on using prompts
\cite{li-etal-2020-unified}, and work by \cite{spanner} to combine the span-prediction and sequence-labeling approaches using the span-prediction approach.

\section{Preliminaries}\label{prelims}
 Let every prediction $p$ for NER be a tuple of the form $$p = (\text{SampleId}, \text{EntityType}, \text{BeginOffset}, \text{EndOffset})$$ which consists of the identifier of the sample/text in which the entity is found, the type of the entity, and the beginning and ending offsets for the entity.

\section{Preprocessing}
For GENIA and NCBI-Disease, each sample is an English sentence. For SocialDisNER, each sample is an entire Spanish tweet. For LivingNER, we use the FLERT\cite{schweter2020flert} approach for document-level NER, in which each Spanish sentence is surrounded by a context of 100 characters to the left and 100 characters to the right.

\begin{table*}[t]
  \centering
\begin{tabular}{|l|l|l|l|l|}
\hline
Dataset & SocialDisNER & LivingNER & Genia & NCBI-Disease \\ \hline
SOTA & \cite{fu-etal-2022-casia-smm4h22} & \cite{vicomtech} & \cite{shen-etal-2022-parallel} & \cite{tian2020improving} \\ \hline
 & 89.1, 90.6, 87.6 & 95.1, 95.8, 94.3 &  81.7, -, - & 90.08, -, - \\ \hline
\texttt{SpanPred} & 90.4, 90.5, 90.4 & 95.7, 95.4, 96.0 & 77.1, 77.0, 77.1 & 89.0, 88.1, 89.9 \\ \hline
\texttt{SEQ} & 88.7, 88.3, 89.1 & 95.0, 94.7, 95.3 & 76.1, 79.8, 72.7 & 88.7, 87.8, 89.5 \\ \hline
\texttt{SeqCRF} & 89.8, 89.6, 90.0 & 95.3, 95.6, 95.0 & 75.7, 79.7, 72.1 & 87.9, 86.2, 89.6 \\ \hline
\texttt{SpanPred} $\cup$ \texttt{SEQ} & 89.0, 86.0, 92.2 & 95.2, 93.4, 97.1 & 77.2, 73.5, 81.4 & 88.2, 84.6, 92.2 \\ \hline
\texttt{SpanPred} x \texttt{SEQ} & 90.2, 93.3, 87.3 & 95.5, 96.9, 94.2 & 75.8, 85.0, 68.5 & \textbf{89.6}, 91.9, 87.4 \\ \hline
\texttt{SpanPred} $\cup$ \texttt{SEQ} $\cup$ \texttt{SeqCRF} & 88.3, 84.1, 93.0 & 94.9, 92.5, 97.4 & 76.4, 71.3, 82.3 & 87.1, 81.4, 93.8 \\ \hline
\texttt{SpanPred} x \texttt{SEQ} x \texttt{SeqCRF} & \textbf{90.8}, 91.2, 90.4 & 95.7, 96.1, 95.4 & 77.1, 81.9, 72.9 & 89.5, 88.8, 90.1 \\ \hline
Meta(\texttt{SpanPred} $\cup$ \texttt{SEQ}) & \underline{90.5}, 89.7, \underline{91.3} & \underline{\textbf{95.7}}, 94.6, \underline{96.9} & \underline{\textbf{78.3}}, 77.4, \underline{79.2} & \underline{89.1}, 86.3, \underline{92.2} \\ \hline
\end{tabular}  

\caption{Performance of all systems on test set on all 4 biomedical datasets. $\cup$ represents the \texttt{Union} combiner and $\text{x}$ represents the \texttt{MajVote} combiner.}
  \label{tab:principle}
\end{table*}

\section{Models} \label{models}

\subsection{\texttt{Seq} model} 

\paragraph{Token Representation Step} 
Given a sentence $\vec{x} = [w_{1},
w_{2}, ..., w_{n}]$ with $n$ tokens, we generate for each token $w_{i}$  
a contextualized embedding $\vec{u}_{i} \in \mathbb{R}^{d}$ 
that corresponds to the last-hidden-layer 
representation of the language model. 
Here, $d$ represents the size of the token embedding.
Importantly, special tokens like \texttt{[CLS]} and \texttt{[SEP]} are also represented. 
We find that the performance can drop significantly(especially for \texttt{SEQ}) if they are not incorporated in the learning process.

XLM-RoBERTa large\cite{xlm} is the multilingual language model that we use for the LivingNER and SocialDisNER spanish tasks.
Inspired by its high performance on the BLURB\cite{gu2021domain} biomedical benchmark, 
we use BioLinkBert large\cite{yasunaga-etal-2022-linkbert} for the NCBI-Disease and GENIA datasets.

\paragraph{Token Classification Step} In this layer, we classify every token
representation into a set of named entity types corresponding to the
BIO(\textit{beginning}, \textit{inside}, \textit{outside}) tagging scheme.
Assuming $\mathbf{\Theta}$ is the set of all named entity types,
then the set of all BIO tags $\mathbf{B}$ is of size $(2 \times |\vec{\Theta}|) + 1$. In other
words, a linear layer maps each token representation $\vec{u}_{i} \in
\mathbb{R}^{d}$ to a prediction $\vec{p}_{i} \in \mathbb{R}^{|\bf{B}|}$, where
$d$ is the length of the token embedding. Finally, the predictions are used to
calculate loss of given sentence $\vec{x}$ with $n$ tokens as follows:
\begin{equation} \text{Loss}(\vec{x}) =
\frac{-1}{n}\sum_{i=1}^{n}\text{log}(\text{Softmax}(\vec{p}_{i})_{y_{i}})
\end{equation} Here $y_{i}$ represents the index of the gold BIO label of the
$i^{th}$ token.

\subsection{\texttt{SeqCRF} Model}
This model is identical to the \texttt{Seq} model except that
we pass the contextualized token representation $\vec{U}$ through a a Linear Chain CRF\cite{lafferty2001conditional} layer.
The CRF layer computes the probabilities of labeling the sequence using the Viterbi algorithm\cite{forney1973viterbi}.
A loss suited to the CRF layer's predictions is then used to train the model. 
We directly use the CRF implementation available in the FLAIR\cite{akbik-etal-2019-flair} framework. The BIO scheme is used for token classification.

\subsection{\texttt{Span} Model}

\paragraph{Token Representation Layer} Same as the token representation layer of the \texttt{Seq} model.

\paragraph{Span Representation Layer} 
Let a span $\vec{s}$ be a tuple $\vec{s} = (b,e)$ where $b$ and $e$ are the
beggining and ending token indices, and $\vec{s}$ represents the text segment
$[w_{b}, w_{b+1}, ..., w_{e}]$ where $w_{i}$ is the $i^{th}$ token.
In this layer, we enumerate \textbf{all possible} spans and then represent 
each span using two token embeddings located at its boundary. 
More precisely, given embeddings  $[\vec{u}_{1}, \vec{u}_{2},
..., \vec{u}_{n}]$ of $n$ tokens, there are $\binom{n}{2} = \frac{n^{2}}{2}$
possible spans, which can be enumerated and represented as the list
$[(0,0), (0,1), ..., (0,n), (1,1), (1,2) ...(1,n),... (n,n)]$. Then we removed all spans that have a length longer than 32 tokens -- this was important to fit the model in GPU memory with a batch size of 4. Finally, as in \cite{lee-etal-2017-end}, each span $s_{i}$ will be represented by
$\vec{v}_{i} = [\vec{u}_{b_{i}};\vec{u}_{e_{i}}]$, a concatenation of the
beginning and ending token embeddings. Hence, the output of this layer is
$\textbf{V} \in \mathbb{R}^{k \times (2 \times d)}$ where $k = \frac{n^{2}}{2}$
and $d$ is length of the token embedding vector.

\paragraph{Span Classification Layer} In this layer, we classify each span
representation with a named entity type. We introduce an additional label
\verb|Neg_Span| which represents the absence of a named entity. Precisely, a
linear layer maps each span representation $\vec{v}_{i} \in \mathbb{R}^{(2
\times d)}$ to a prediction $\vec{p}_{i} \in \mathbb{R}^{|\Omega|}$, where
$\Omega$ is the set of all named entity types(including \verb|Neg_Span|) and
$d$ is the size of the token embedding. Finally, the predictions are used to
calculate loss of given sentence $\vec{x}$ with $l$ possible spans as follows:
\begin{equation} \text{Loss}(\vec{x}) =
\frac{-1}{l}\sum_{i=1}^{l}\text{log}(\text{Softmax}(\vec{p}_{i})_{y_{i}})
\end{equation} Here $y_{i}$ represents the index of the gold label of the
$i^{th}$ span.

\subsection{\texttt{Union} combiner model}
This model doesn't learn weights. For a given list $P_{0}, P_{1}, ..., P_{n}$ where $P_{i}$ is the set of predictions(as defined in \autoref{prelims}) made by the $i^{th}$ NER model and $n$ is the total number of models, it returns the set $P_{1} \cup P_{2} \cup ... P_{n}$.   

\subsection{\texttt{MajVote} combiner model}
This model doesn't learn weights. This is the classic majority voting combiner model. Precisely,
when given a list $P_{0}, P_{1}, ..., P_{n}$ where $P_{i}$ is the set of predictions(as defined in \autoref{prelims}) made by the $i^{th}$ NER model and $n$ is the total number of models, it returns a set which only includes predictions in $P_{1} \cup P_{2} \cup ... P_{n}$ that have been predicted by more that $\lfloor\frac{n}{2}\rfloor$ models.

\subsection{\texttt{Meta} combiner model}
The job of meta is simple : "Learn to tell if a prediction made by \texttt{SEQ} or \texttt{SpanPred} is a mistake or not". In other words, \texttt{Meta} looks at a prediction made by \texttt{SEQ} or \texttt{SpanPred} on the \textit{validation set} and learns to classify the prediction as being either "correct" or "incorrect". "correct" means that the prediction is a good prediction, and that it should not be removed. "incorrect" means that the prediction should be removed. In other words, if $P_{\text{SEQ}}$ is the set of all predictions of the \texttt{SEQ} and $P_{\text{Span}}$ is the set of all predictions of \texttt{SpanPred}, then \texttt{Meta} acts as (and learns to be) a filter for $P_{\text{Span}} \cup P_{\text{SEQ}}$. During evaluation, \texttt{Meta} filters $P_{\text{Span}} \cup P_{\text{SEQ}}$, generating a final set of predictions.

\par We borrow the idea of using markers made with special tokens \cite{baldini-soares-etal-2019-matching, xiao-etal-2020-denoising, ye-etal-2022-packed} which, intuitively, help models "focus their attention on the span-of-interest". In other words, by introducing special tokens(which act as markers) like \texttt{[e]} and \texttt{[/e]} in the language model's vocabulary, and then surrounding the span-of-interest with them, one can help the model "focus" of the span of interest while making some prediction. In Meta's case, the markers are supposed to help locate/identify the entities predicted by \texttt{SEQ} or \texttt{SpanPred} in raw text. See \autoref{meta-example} for an example input prediction with markers highlighting the entity.\par 

We also borrow the idea of prompting\cite{li-etal-2020-unified}, which involves pre-pending some text(prompt) to the original input text with the goal of priming(or aiding) a model's decision making with a useful bias. In particular, every input to \texttt{Meta} includes the type of the predicted entity as prompt. Intuitively, this helps \texttt{Meta}  recognize the type of the entity it is dealing with. See \autoref{meta-example} for an example of prompting with the entity type "disease".\par

 Note that prompting and special markers are \textit{only} used to prepare the training data for \texttt{Meta} using the predictions of \texttt{SEQ} and \texttt{SpanPred} on the validation set. \texttt{Meta} itself is a simple binary classification neural model. Just like \texttt{SEQ}, \texttt{SeqCRF} and \texttt{SpanPred}, it first creates contextualized token representations from raw input using the appropriate language model(XLM-RoBERTa or BioLinkBERT) and then classifies the pooler token(\texttt{[CLS]} or \texttt{[s]}) representation using a linear layer. As in \texttt{SpanPred} and \texttt{SEQ}, cross-entropy loss is used to train the model.\par

Because \texttt{META} acts as a "filter"(it allows certain predictions and disallows others), it \textit{cannot} improve recall -- it can only improve precision. Ideally, \texttt{Meta} will learn the true nature of the mistakes  that \texttt{SEQ} and \texttt{SpanPred} make and remove all false positives, resulting in a perfect precision score of 100 and no drop in recall. 

\paragraph{Preparing the training data for \texttt{Meta}:}
\textit{all} predictions(with "correct" and "incorrect" labels) on the validation set for \textit{all} 20 epochs by \textit{both} \texttt{SEQ} and \texttt{SpanPRED}, and \textit{all} gold predictions(that only have "correct" labels) from the \textit{original} training data make up the training set for \texttt{Meta}. We hold out 15 percent of \texttt{Meta}'s training set for validation. Note that we incorporate the predictions of \texttt{SpanPred} and \texttt{SEQ} from earlier epochs because the fully trained high-performing models don't make that many mistakes(which META needs for its learning). As expected, the test set is not touched while training \texttt{Meta}. During evaluation, \texttt{Meta} filters the predictions made by \texttt{SEQ} and \texttt{SpanPred} on the test set.
 
\subsection{Meta input example}\label{meta-example}
Assume the example sentence "\texttt{Bob has HIV and flu.}" and the task of identifying diseases. Now assume that \texttt{SEQ} predicted \newline (id, \textbf{disease}, 8, 11) (see \autoref{prelims} for the definition of prediction) and correctly identified the disease "HIV" in the input. Then, the input to meta will be the the text \texttt{"\textbf{disease} Bob has [e] HIV [/e] and flu"} and the associated gold label of \texttt{correct}. Prompting with \textbf{disease} informs \texttt{Meta} that it is dealing with a prediction representing a disease. \texttt{Meta} has to make a judgement on whether the prediction is correct or not. 

\subsection{Training and Optimization} 
Both XLM RoBERTa large\cite{xlm} and BioLinkBERT large\cite{yasunaga-etal-2022-linkbert} are fine-tuned on the training data using the Adafactor\cite{adafactor} optimizer with a learning rate of \verb|1e-5|(see \href{https://github.com/flyingmothman/bionlp/blob/d61b02593711b43b5d0f00f0c6ed62fb7685adf3/utils/training.py#L13-L20}{code}) and a batch size of 4 for \textit{all 4 datasets}. Specifically, we used \href{https://huggingface.co/docs/transformers/main_classes/optimizer_schedules#transformers.Adafactor}{the implementation of Adafactor} available on HuggingFace\cite{hugging}. It was not possible for us to use the same learning rate and batch size for every dataset with Adam\cite{adam} because we noticed it was prone to over-fitting(and then collapsing) mid-training on LivingNER, NCBI-Disease, and GENIA -- batch-size had to be increased to avoid over-fitting. Moreover, we found that \texttt{SEQ}, \texttt{SeqCRF}, and \texttt{SpanPred} converged to better solutions with Adafactor on all datasets. However, we found that \texttt{Meta} consistently converged to better solutions on the NCBI disease dataset using Adam. \par The best model is selected using early stopping with a patience(in terms of epochs) of 5. 

\section{Evaluation Methodology}
All tasks evaluate systems using the strict(no partial matching) Micro F1, Precision and Recall. For SocialDisNER, \textit{all} systems were submitted to the corresponding CodaLab\cite{pavao2022codalab} competition website for evaluation. For LivingNER, \textit{all} our systems have been evaluated using the \href{https://temu.bsc.es/livingner/2022/01/28/evaluation/}{official evaluation script} that the organizers made available. For Genia and NCBI-Disease, we unfortunately couldn't find official CodaLab websites, so we had to use our own script, which can be inspected \href{https://github.com/flyingmothman/bionlp}{here}.

\section{Analysis of Results}
Note that among the 3 models, \texttt{SpanPred} consistently outperforms the other two on all datasets. This is anticipated on tasks with overlapping entities like LivingNER and GENIA(because \texttt{SEQ} and \texttt{SeqCRF} cannot represent them), but not on "flat" NER tasks like SocialDisNER and NCBI-Disease. \par
Note that any system resulting from a \texttt{Union} combination should have higher recall than any of the involved systems because a set union operation is incapable of removing a correct prediction (the set of false negatives can only shrink with more systems). Also, the resulting system's precision cannot be higher than the highest precision observed in any sub-system.   \autoref{tab:principle} adheres to both of these expectations. On the other hand, a system resulting from a \texttt{MajVote} combiner is \textit{inclined} to have higher precision when the systems being combined are diverse and comparable because -- intuitively --  \texttt{MajVote} can be a more "picky" system (only allowing a prediction if it has been voted on by several). In \autoref{tab:principle}, note that both \texttt{SpanPred\textbf{x}SEQ} and \texttt{SpanPred\textbf{x}SEQ\textbf{x}CRF} consistently boost precision across all datasets. Also note that the best \texttt{MajVote} systems significantly outperform all other systems on precision while maintaining the highest F1 on all datasets except Genia, where \texttt{Meta} outperforms all other systems on F1 for the first(and last) time. Also on Genia is the only time when a \texttt{Union} model ($\texttt{SpanPred} \cup \texttt{SEQ}$) outperforms the \texttt{MajVote} models due to a significant boost in recall. Finally, note how \texttt{Meta}, across all datasets, outperforms \texttt{SpanPred}, \texttt{SEQ}, and \texttt{SeqCRF} models on Recall and delivers an F1 that is at least as high as any of the three models.\par

\section{Conclusion}
Our implementation(\href{https://github.com/flyingmothman/bionlp}{code available}) of \texttt{CRF} and \texttt{SpanPred}, two simple models, improves the state of the art on LivingNER and SocialDisNER datasets. We used two simple approaches called \texttt{Union} and \texttt{MajVote} to combine the NER models' predictions and studied the results. \texttt{MajVote} on the three NER models seems to be effective at generating systems with high precision and high F1. While \texttt{Union} can generate systems with higher recall, it is only at the cost of F1 due to a significant drop in precision. \texttt{Meta} seems to be effective at alleviating \texttt{Union}'s issue, generating systems with both high recall and high F1.


\bibliography{anthology,custom} \bibliographystyle{acl_natbib}



\end{document}